\documentclass[letterpaper, 10 pt, conference]{ieeeconf}  

\IEEEoverridecommandlockouts                              

\usepackage{graphics} 
\usepackage{epsfig} 
\usepackage{mathptmx} 
\usepackage{times} 
\usepackage{amsmath} 
\usepackage{amssymb}  
\usepackage{xcolor}
\usepackage{comment}
\usepackage{mwe}
\usepackage{caption}
\usepackage{graphicx}
\usepackage{pbox}
\usepackage{tabularx,booktabs}
\usepackage{float} 
\usepackage{subfigure}
\usepackage{comment}
\usepackage{soul}

\usepackage{fancyhdr}
\fancypagestyle{firstpage}{%
    \chead{This paper has been accepted for publication at the 2023 International Symposium on Medical Robotics.}
    }

\title{\LARGE \bf
Towards Biomechanics-Aware Design of a Steerable  Drilling Robot \\for Spinal Fixation Procedures with Flexible Pedicle Screws
}

\author{Susheela Sharma*, Yuewan Sun*, Sarah Go,  Jordan P. Amadio, Mohsen Khadem, Amir Hossein Eskandari, \\ and Farshid Alambeigi, \IEEEmembership{Member, IEEE}
\thanks{*These authors contributed equally}
\thanks{**This work is supported by the National Institute Of Biomedical Imaging and Bioengineering of the National Institutes of Health under Award Number R21EB030796.}
\thanks{S.~Sharma, Y.~Sun, S.~Go, and F.~Alambeigi are with the Walker Department of Mechanical Engineering and Texas Robotics at the University of Texas at Austin, Austin, TX, 78712, USA. Email: \{sheela.sharma, yuewansun, sarah.go\}@utexas.edu,  farshid.alambeigi
@austin.utexas.edu}.
\thanks{J.~P.~ Amadio is with the Department of Neurosurgery, The University of Texas Dell Medical School, TX, 78712. }
\thanks{M.~Khadem is with the School of Informatics, University of Edinburgh, UK.}
\thanks{A.~H.~Eskandari is with Institut de recherche Robert Sauvé en santé et en sécurité du travail, Montréal, Canada.}
}

\begin{document}

\maketitle
\thispagestyle{firstpage}
\pagestyle{empty}


\begin{abstract}
Towards reducing the failure rate of spinal fixation surgical procedures in osteoporotic patients,  we propose a unique biomechanically-aware framework for the design of a novel concentric tube steerable drilling robot (CT-SDR). The proposed framework leverages a patient-specific finite element (FE) biomechanics model developed based on Quantitative Computed Tomography
(QCT) scans of patient's vertebra to calculate a biomechanically-optimal and feasible drilling and implantation trajectory. The FE output is then used as a design requirement for the design and evaluation of the CT-SDR.  
Providing a balance between the necessary flexibility to create curved optimal trajectories obtained by the FE module with the required strength to not buckle during drilling through a hard simulated bone material, we showed that the CT-SDR can reliably recreate this drilling trajectory  with  errors between 1.7-2.2\%.

\end{abstract}

\section{INTRODUCTION}\label{Intro}
Osteoporosis is responsible for an estimated two million broken bones in the United States every year \cite{burge2007incidence}.
The most common type of osteoporotic fracture occurs in the spine, with more than 1.4 million global incidence in men and women within this age category \cite{johnell2006estimate}. To eliminate painful motion caused by these fractures and  restore stability of the spine, spinal fixation  surgical procedures are commonly performed by fusing and locking the involved vertebrae together into a solid bone. 
While this procedure is well established and widely practiced around the world, for patients suffering from osteoporosis, this  fixation technique using a rigid screw is often insufficient and the fixation may fail \cite{weiser2017insufficient}. This failure can be mainly attributed to the lack of dexterity and rigidity of the implants and instruments used by the surgeons in the operating room. Of note, this critical limitation together with the complex anatomy of vertebrae constrain the surgeons to linear drilling trajectories as a pilot hole for implantation of  rigid pedicle screws within the vertebral body. Also, this limitation increases the risk of implanting the screw within the lower bone mineral density (BMD) regions of the vertebral body that may results in a fixation failure.  Aside from non-optimal instruments and implants, the  drilling  and, therefore, screw implantation  trajectory in spinal fixation procedures  often is  not optimally selected based on the biomechanical analysis-- and particularly calculating the stress and strain distribution along the implanted screw \cite{bakhtiarinejad2019biomechanical}. Nevertheless, a successful spinal fixation procedure demands a biomechanically-aware design of flexible instruments and implant to collectively address the above-mentioned issues.

\begin{figure}[t]
	\centering
	\includegraphics[width=1.0\linewidth]{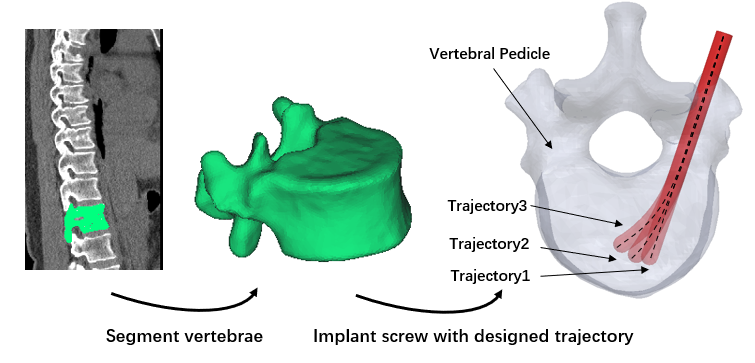}
	\caption{Procedure of building up an FE model from QCT scans. Figure also shows the considered desired drilling trajectories.}
	\label{fig:concept}
	\vspace{-8 pt}
\end{figure}

Despite conventional spinal fixation devices, flexible instruments and implants (e.g., \cite{alambeigi2018inroads,alambeigi2019use,alambeigi2016toward}) enable surgeons to drill and place implants through  areas within the vertebral body with a higher BMD index, and therefore, potentially reduce the risk of fixation failure. In an effort to overcome  the aforementioned dexterity limitations and reduce the failure rates of spinal fixation procedures, several robotic drilling systems have been developed in the literature. For example, Alambeigi et al. \cite{alambeigi2019use,alambeigi2017curved,alambeigi2020steerable} designed a tendon-driven curved drilling robot for core decompression of femoral head osteonecrosis. While the proposed system was capable of drilling relatively smooth, curved holes, the system had several limitations, such as the lengthy 5-9 min drilling time and a maximum path length of 35 mm. These two limitations namely stem from a lack of structural rigidity present in tendon-actuated flexible robots, leaving the system prone to buckling and drilling failures at high insertion speeds. 
Expanding on this work, Ma et al. \cite{ma2021active} developed a handheld version of the described robotic system. This system suffers from the same limitations present in its predecessor, in that the stiffness and bending behaviour are limited by the maximum loading capacity of the tendons and the material being drilled through.  In a more recent effort, also
Wang et al. \cite{wang2021design}  proposed a tendon-driven articulated surgical drill for spinal fusion applications. The proposed design included an articulated wrist that enables drilling in linear trajectories connected by a corner between the two intersecting pathway. Of note, this limits the ability to drill a smooth curvilinear trajectory within vertebra for implantation of a flexible implant for the fixation procedure. Also, a review of these literature demonstrates that  all of these steerable drilling systems  have neglected considering the effect of biomechanics  during the design and evaluation of the  robotic system and defining the drilling trajectory.

To collectively address the above-mentioned limitations, in this paper, we propose a novel biomechanically-aware procedure to design, fabricate, and evaluate an easy-to-control Concentric Tube Steerable Drilling Robot (CT-SDR) for enhancing fusion stability within osteoporotic vertebrae. Inspired by concentric tube robots (e.g., \cite{dupont2009design,webster2010design}) and designed to match the results given from a novel finite element (FE) analysis framework, the CT-SDR provides quick, reliable, and accurate access to areas of high BMD for flexible screw implantation by following a desired trajectory obtained from FE framework shown in Fig. \ref{fig:concept}. 


\section{Design Requirements for the CT-SDR}

To effectively create a solution to address the aforementioned limitations discussed in section \ref{Intro}, a  biomechanically-aware steerable drilling robot (SDR) needs to be designed to  (1)  guide a flexible drilling instrument and intuitively  drill a desired straight/curved  trajectory in a clinically-acceptable amount of time,  (2) provide adequate structural stiffness to prevent its bending/buckling and ensure a predictable and repeatable drilled trajectories, and most importantly (3) re-create the desired trajectory determined through a patient-specific FE analysis. Of note, the  last requirement is the critical step for a  ``biomechanically-aware approach" towards the design of a SDR for a spinal fixation procedure.

 To summarize, the proposed  biomechanically-aware approach, firstly uses the Quantitative Computed Tomography (QCT) scans of  patients to obtain their spatial BMD  distribution for the FE analysis and calculating the stress and strain distribution along an implanted screw in a straight or curved trajectory. Next, using this FE analysis, a desired linear/curved drilling trajectory is determined based on appropriate biomechanical metrics. This desired trajectory is then used for the design and evaluation of a SDR. The SDR should be able to successfully provide adequate structural stiffness and workspace to reliably drill through a hard tissue while followings this desired trajectory.   The following sections describes these steps in details.


\section{FINITE ELEMENT MODEL BUILDUP}\label{FEA Model}
\subsubsection{Image analysis}
\begin{figure}[t]
\centering 
\subfigure[Trajectory 1]{
\includegraphics[width=2.6cm,height = 2.8cm]{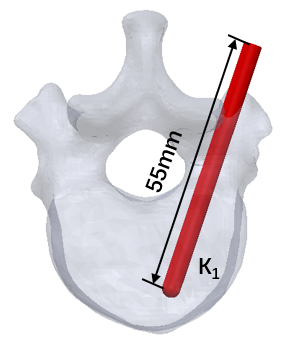}}\subfigure[Trajectory 2]{
\includegraphics[width=2.5cm,height = 2.8cm]{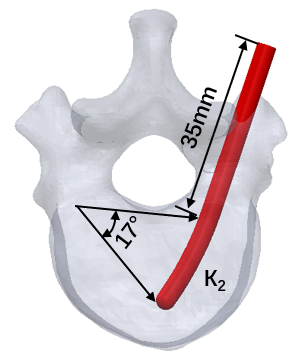}}\subfigure[Trajectory 3]{
\includegraphics[width=2.5cm,height = 2.8cm]{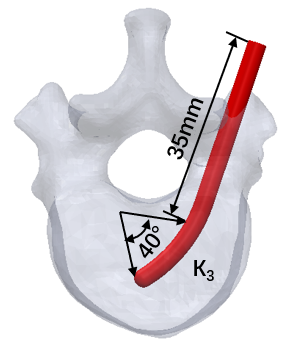}}
\caption{The three designed trajectories analyzed in this paper. Curvature $\kappa_1$  is    infinite. Curvature $\kappa_2$ is 1/0.0695 m=14.388 m$^{-1}$ and   curvature $\kappa_3$ is 1/0.035 m=28.571 m$^{-1}$.}
\label{trajectory design}
\end{figure}

Quantitative Computed Tomography (QCT) images of a real patient (male, 52 years) were obtained using a calibration phantom \cite{KALENDER199583} to build up the FE model of a lumbar T12 vertebrae. The protocol of image acquisition was approved by the review board at the University of Texas at Austin (IRB ID: STUDY00000519). The QCT scan settings included: a 0.625 mm slice interval, 22.6 cm field of scan, and 512 $\times$ 512 pixel resolution. To start the FE analysis, DICOM images were first imported into MIMICS software (Materialise, Leuven, Belgium), and segmented with a threshold of 100 HU before being smoothed  \cite{MASTMEYER2006560}. The 3D surface model of the vertebra was then exported as an STL into 3D CAD modelling software (SolidWorks, Dassault Systemes S.A., USA). Subsequently, as shown in Fig. \ref{trajectory design}, for our analysis in this paper, three different trajectories were designed and implanted with their corresponding simplified screws following three Trajectories called Trajectory 1-3 throughout the paper. As can be seen, Trajectory 1  was a straight reference trajectory for a rigid pedicle screw while  Trajectory 2 and 3 were curved trajectories for an implanted flexible pedicle screws curved at 25 mm from the posterior end with curvatures of 14.388 m$^{-1}$ and 28.571 m$^{-1}$, respectively (Fig. \ref{trajectory design}). After designing trajectories, a screw with simplified geometry (D = 2.5 mm, L = 55 mm) was implanted into the considered vertebral trajectories. Next, the designed geometries were imported into Hypermesh software (Altair Engineering, Troy, MI, USA) and meshed using 4-node tetrahedral elements with average element size of 0.8 mm \cite{article}. To increase simulation accuracy, common node method was used on the interface of the vertebra and screws. 

To simulate the heterogeneity of the bone material distribution, the elastic modulus of each bone element was estimated directly from the QCT data by (i) converting the Hounsfield value of element nodes to their density values, $\rho=1.122*HU+47$ \cite{jain_biomechanical_2018} and then (ii) converting them into the corresponding elastic modulus (cancellous bone: $E=0.63\rho^{1.35}$; cortical bone: $E=1.89\rho^{1.35}$) \cite{inproceedings} and lastly (iii) taking the average of nodes density. The Hounsfield value used to separate the cortical and the cancellous bone was set at 1800 \cite{rho1995relations}. The average values of the elastic modulus for the cancellous and cortical bone were roughly 2500 and 15,000 MPa, respectively. The material of screw was simulated as stainless steel with the elastic modulus set to 200 GPa \cite{Lim1996BiomechanicsOT}. All materials were assumed to be linear and isotropic, with the Poisson ratio set to 0.3 \cite{skalli1993biomechanical}.

\begin{figure}[t]
	\centering
	\includegraphics[scale=0.4]{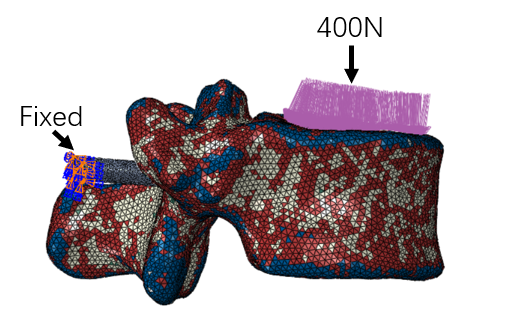}
	\caption{Loading condition on finite element model.}
	\label{fig:load}
\end{figure}

\subsubsection{FE model and boundary conditions}
The meshed geometries of the vertebra and screws were imported into the Abaqus software (Dassault Systemes S.A., USA). Loading and boundary conditions were taken from literature in-vitro studies  \cite{PMID:9051885} with a distributed downward force of 400 N \cite{PMID:11145808} applied on the superior endplate of the vertebral body while the end of the pedicle screw was clamped, as shown in Fig. \ref{fig:load}. The interface was modeled as a surface-to-surface contact interaction with friction coefficient of 0.2 \cite{chen2003biomechanical}. The biomechanical performance of the designed trajectories were investigated using the maximum stress and strain criteria in cancellous bone where the screw loosening and pulling out happens \cite{rometsch2020screw}.

\vspace{-4 pt}
\subsection{FE Simulation and Results}\label{Traj Results}
\begin{figure}[t]
\centering 
\subfigure[]{
\includegraphics[width=2.5cm,height = 2.0cm]{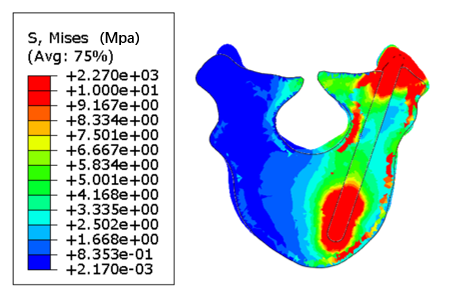}}\subfigure[]{
\includegraphics[width=2.5cm,height = 2.0cm]{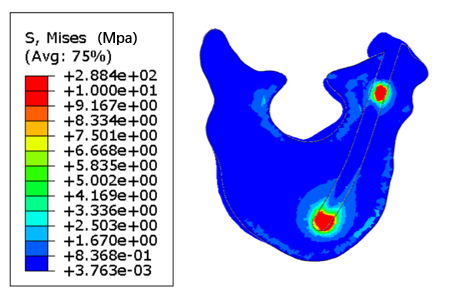}}\subfigure[]{
\includegraphics[width=2.5cm,height = 2.0cm]{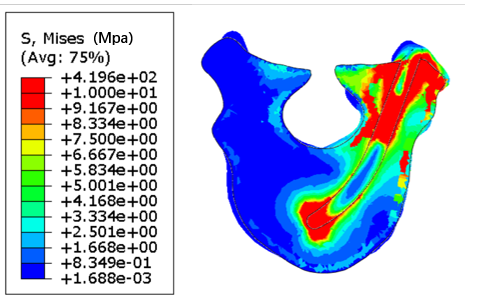}}
\caption{Stress distribution  of the three considered trajectories.}
\label{stress results}
\vspace{-5 pt}
\end{figure}

\begin{figure}[t]
\centering 
\subfigure[]{
\includegraphics[width=2.5cm,height = 2.0cm]{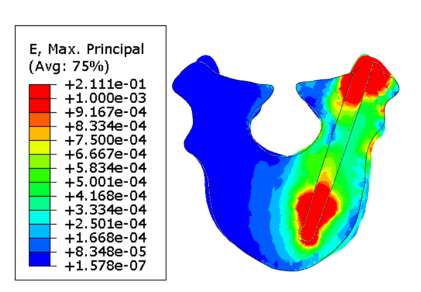}}\subfigure[]{
\includegraphics[width=2.5cm,height = 2.0cm]{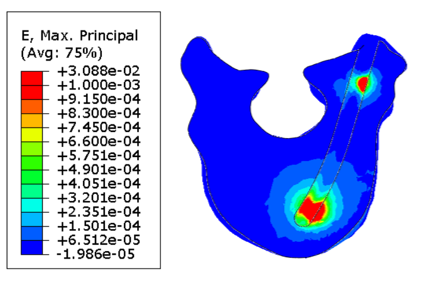}}\subfigure[]{
\includegraphics[width=2.5cm,height = 2.0cm]{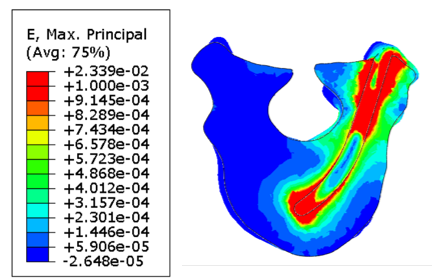}}
\caption{Strain distribution  of the three considered trajectories.}
\label{strain results}
\vspace{-5 pt}
\end{figure}

 To better illustrate the stress distribution in the cancellous bone, the upper limit of the legend was set to 10 MPa in Fig. \ref{stress results}. The results shows that besides the high stress region in the pedicle region, there was a high stress region in cancellous bone for all three trajectories. However, the maximum stress of Trajectory 2 was significantly lower than the Trajectory 1 and  Trajectory 3 (Fig. \ref{stress results}(b)). More specifically, the maximum stress in cancellous bone region were calculated as 1.01 GPa, 0.20 GPa, and 0.34 GPa for Trajectory 1- Trajectory 3, respectively. Moreover, to investigate the failure of cancellous bone, strain distribution of the trajectory section was illustrated in Fig. \ref{strain results}. The maximum strain was set to 1e-3. Although, all three trajectories showed high strain regions in cancellous bone, Trajectory 2 (Fig. \ref{strain results}(b)) had least maximum strain area in comparison with the other trajectories which implies lower potential of cancellous bone failure. The strain for the three trajectories were obtained as 9.11e-2, 2.05e-2, and 2.129e-2. Based on this analysis, therefore, we considered Trajectory 2 as a biomechanically-aware desired trajectory for the design of the CT-SDR. 


\section{Design of the CT-SDR}\label{CT-SDR Design}
\begin{figure*}[!t]
	\centering
	   \includegraphics [scale=.43]{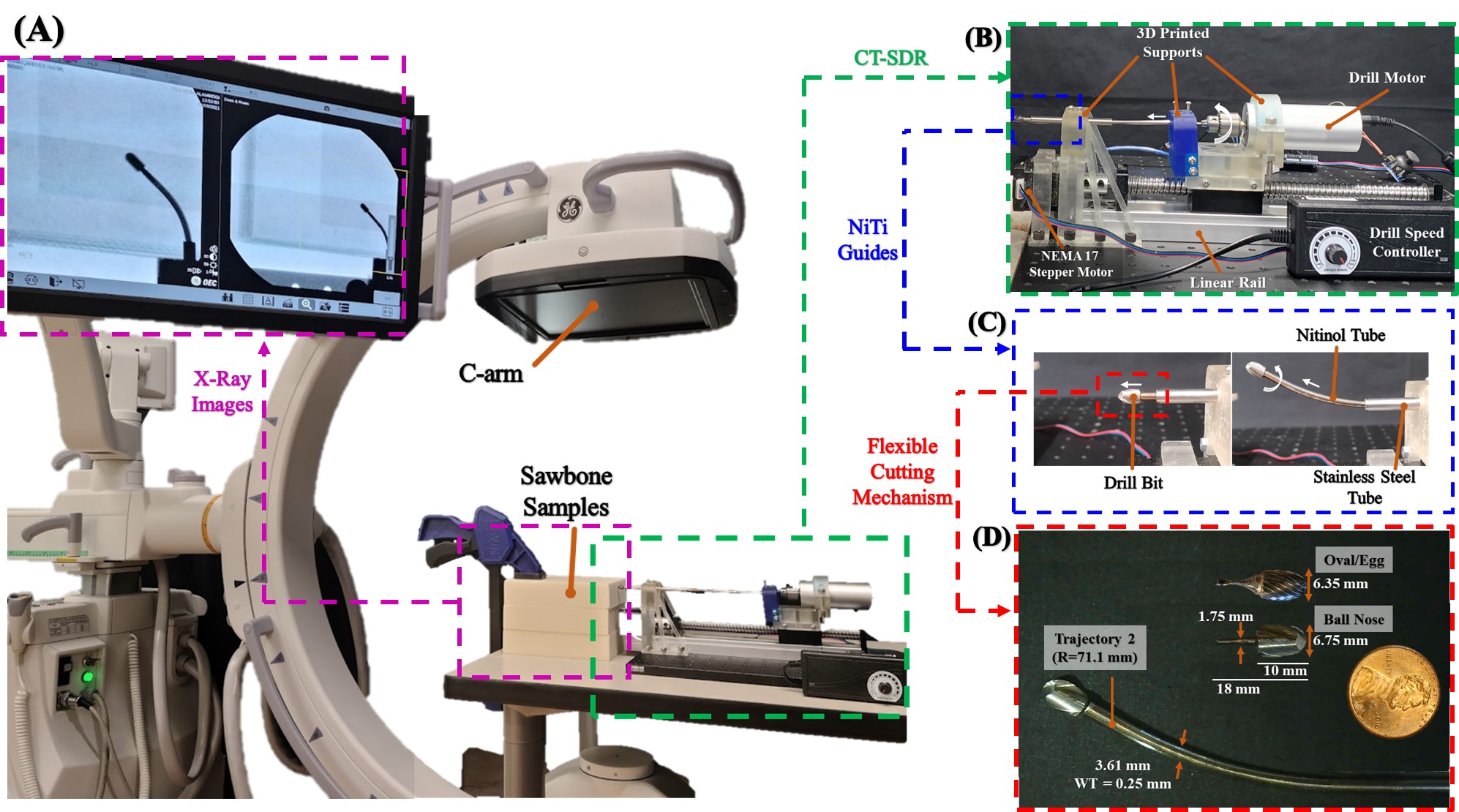}
		\caption{Breakdown of the CT-SDR design along with details of the experimental set-up. (A) shows the C-arm X-ray machine, its position during testing and the view shown during experimentation. (B) displays the CT-SDR with labels of critical parts of the actuation unit. (C) shows a closer view of a NiTi guide positioning the drill tip in free space. (D) shows the NiTi guide used in this paper and the 2 drill bits used in our experiments, crucial to the flexible cutting mechanism.}
		\label{fig:set-up}
	\end{figure*}

As mentioned, to meet the mentioned design requirements and inspired by the concentric tube robots (e.g., \cite{dupont2009design,webster2010design,burgner2015continuum}), we propose a CT-SDR.  As shown in Fig. \ref{fig:set-up}, the proposed CT-SDR includes three main components, (i) flexible components to guide the drill through the optimal Trajectory 2 obtained in section \ref{Traj Results} while providing the strength to resist buckling; (ii) a flexible shaft to transmit torque from the drill motor to the drill's tip; and (iii) an actuation system to allow the surgeon to reliably control the system. The following sections will discuss these three components and how they meet the design requirements set forth in this paper.

\vspace{-4 pt}
\subsection{Concentric Guiding Tubes}
As shown in Fig. \ref{fig:set-up},  to provide simultaneous flexibility and structural stiffness, we propose utilizing two concentric tubes including one rigid stainless steel to function as the outer tube and one pre-curved nitinol to act as the inner flexible tube. Notably, nitinol is a superelastic, biocompatible, shape memory alloy comprised of nickel and titanium which is being used with a growing frequency for medical technologies and instruments. Within the design, the stainless steel is static, while the nitinol tube is actuated through it. When at its starting position, the nitinol is constrained to a straight trajectory matching that of the stainless steel tube it is held within. As the nitinol tube is actuated forward and out of its stainless steel counterpart, the tube returns to its pre-curved shape, and guides any flexible tools housed within it to do the same.
To program the biomechanically-optimal drilling Trajectory 2 obtained in section \ref{Traj Results}, the 70 mm long NiTi tube (Euroflex GmbH, Germany) was heat treated in a furnace, and then nested within a straight stainless steel tube (89895K421, McMaster-Carr) with 80 mm length, outer diameter of 3.175 mm, and 1.25 mm wall thickness. The outer diameter of the nitinol tube was 3.61 mm with the wall thickness 0.25 mm (Euroflex GmbH, Germany).  This piece was rigidly attached to the linear stage of the actuation unit to control the insertion of the CT-SDR's tip. Fig.  \ref{fig:set-up} shows the heat treated nitinol tube based on the obtained desired trajectory and curvature in section \ref{Traj Results}. Of note,  due to spring-back in the heat treated NiTi material, the resulting guide had a radius of 71.1 mm ($\kappa = 14.065 m^{-1}$), which is slightly less than the desired biomechanically-obtained curvature $\kappa = 14.388 m^{-1}$, shown in Fig. \ref{trajectory design}. 

\vspace{-4 pt}
\subsection{Flexible Shaft}
Each flexible cutting tool is comprised of 3 parts: a straight tube/rod to be grasped by the actuation unit's motors, a flexible torque coil to transmit power along the curved drilling trajectories, and a rigid drill tip capable of cutting in multiple directions. The pieces were rigidly attached together using an epoxy adhesive (1813A243, McMaster-Carr). 
Two different tools were assembled to evaluate the influence of different factors on the performance of the CT-SDR.
The first tool comprised of a stainless steel rod of diameter 1.56 mm (888915K11, McMaster-Carr), a 70 mm length torque coil (Asahi Intec. USA, Inc.), and a 6.35 mm diameter carbide oval/egg head bur (42955A35, McMaster-Carr) modified after purchase to have a shank of 1.5-2 mm diameter and 8 mm in length with a cutting length of 10 mm (Fig. \ref{fig:set-up}). The second tool different from the first in that is utilized a brass tube of diameter 1.56 mm (8859K231, McMaster-Carr), and a 6.75 mm diameter carbide ball nose end mill (8878A42, McMaster-Carr). 

\vspace{-4 pt}
\subsection{Actuation Unit}

The actuation unit of the CT-SDR consists of a linear stage with a linear ball screw (B085TG12D1, Amazon) controlled by a NEMA 17 stepper motor to provide the insertion DoF of the drill's tip, a mini electric hand drill (B075SZZN4J, Amazon) to control the drill rotational/cutting speed, and 3D resin printed mounts and supports. An Arduino R3 micro-controller board and custom-written program controlled the stepper motor's motion and thus the linear speed of the stage. The provided controller for the electric hand drill controlled the drill bit's rotational speed. Of key importance in the CT-SDR's design was the alignment of the concentric tubes with one another and the force required to actuate the nitinol tube through the constraining stainless steel. The NEMA 17 stepper motor allowed for a compact design that met the listed requirements while allowing for precise control of the system throughout testing. To ensure concentricity, the 3D printed supports rigidly held various parts of the design aligned 
and stabilized the entire system by mounting to an optical board.

\section{CT-SDR Evaluation Experiments and Results}
\subsection{Experimental Set-Up}
We designed the test set-up shown in Fig. \ref{fig:set-up} to best replicate the procedure a surgeon would follow in the operating room A Sawbone biomechanical bone model phantom (block 10 PCF, Pacific Research Laboratories, USA) was secured in front of the CT-SDR as if a surgeon had aligned it with the entry point of the patient's pedicle. Notably, PCF 10 was selected as it simulates bone with medium osteoporosis \cite{ccetin2021experimental}. The test then began with the tip of the CT-SDR cutting a trajectory through the test sample, while the test's progress was monitored with a C-arm X-ray machine (OEC One CFD, GE Healthcare) set up perpendicular to the cutting plane of the CT-SDR. As the CT-SDR was actuated through the sample, the C-arm was able to show the location of the drill tip, steering cannula, and cutting trajectory. This allowed us to monitor the test in real time and presented a second avenue of data collection/analysis after the test's conclusion. An additional camera was set up separate from the other components to track cutting time and review the drill's performance after the test's conclusion.

\vspace{-4 pt}
\subsection{Drilling Experiments}
The experiments run with the CT-SDR were designed to evaluate the system's ability to match the requirements set forth in section \ref{CT-SDR Design} and the Trajectory 2 evaluated in section \ref{Traj Results}. The repeatability and predictability of the cutting trajectory was of particular interest during these experiments. During initial testing with the CT-SDR, and in line with similar studies performed on compliant drilling devices \cite{alambeigi2017curved, ma2021active,wang2021design}, several factors were determined to have a potential effect on the system during a procedure: (1) the insertion speed of the guiding tube, (2) the rotational speed of the drill tip, (3) the geometry of the drill tip. To evaluate these factors, two different experiments were performed and thoroughly evaluated with two different flexible instruments.



\begin{figure}[!t] 
		\centering 
		\includegraphics[scale=0.5]{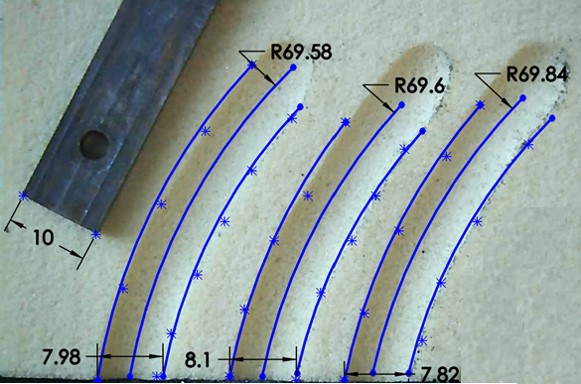}
		\caption{Experimental Result Analysis. View of 3D CAD measurements of mid-line cross section of tests run with the 0.85 mm/s insertion speed and a drill speed of 8250 rpm. Also shown is the reference measurement used to calibrate the 3D scan.}
		\label{fig:Sawbone Results}
		\vspace{-8 pt}
	\end{figure}

After initial testings with the CT-SDR, we determined that a range of insertion speeds would be evaluated to identify their effect on the drill's performance: 0.5, 0.85, 1.25 mm/s. Along with three different rotational speeds of the drill's tip: 6000, 8250, 10,600 rpm. Starting with the oval head drill tip, a total of 30 tests were performed; 5 different tests for the above ranges each repeated 3 times. In these tests, the cutting plane was held fixed, parallel to the optical table on which the CT-SDR was mounted. After the conclusion of these tests, the ball nose end mill was evaluated with insertion speeds of 0.85 and 1.25 mm/s, and a drilling speed of 8250 rpm.  In each trial, the CT-SDR was advanced for the full length of the heat treated NiTi trajectory while the C-arm (intermittently) and the external camera (continuously) recorded the test's progress.
After the completion of testing, the Sawbone test samples were cut along the midline of the entrance holes and photographed using a digital microscope (Jiusion, China). The images were then analyzed using a 3D CAD Software (SolidWorks, Dassault Systèmes) to determine the radius of curvature for each test and how they compare to the actual and ideal trajectories of the NiTi guides. Fig. \ref{fig:Sawbone Results} shows the view of the cross sections along with the 3D software measurements.

\begin{figure}[!t] 
	\centering 
		\includegraphics[scale=0.4]{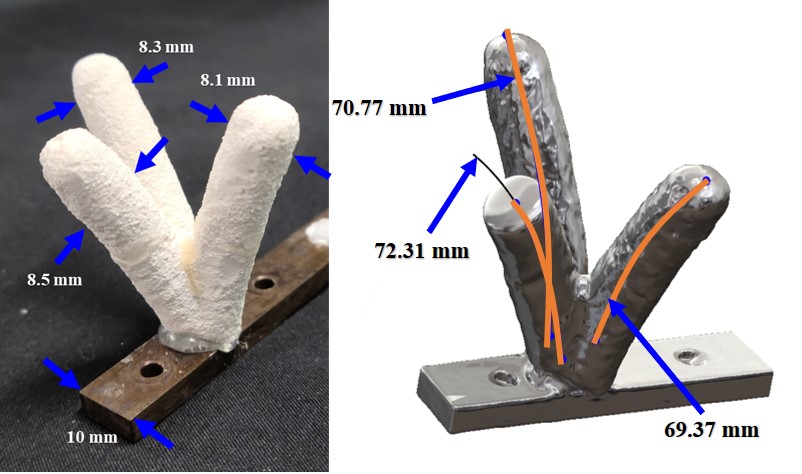}
		\caption{(Left) Plaster mold of a branch test, along with the widths of each path drilled and the reference measurement used to calibrate the software. (Right) 3D Rendered view of the same branch test completed with Trajectory 2, along with the measured mid-lines.}
	\label{fig:3D Branches}
	\vspace{-4 pt}
\end{figure}

To explore further applications of the CT-SDR, we decided to test the system's ability to reach different goal points from a single entry location, or out of plane branch drilling. This would minimize any unnecessary material removal and bone weakening caused by multiple entry points and extra drillings \cite{alambeigi2017curved}. Based on the results from the previous drilling experiments, these tests were performed with an insertion speed of 0.85 mm/s, a rotational speed of 8250 rpm, and with both drilling instruments. The C-arm used in the Trajectory 2 cutting experiments is useful primarily for 2D images/views across a test sample, but is not ideal for visualization in 3D space. Therefore, to create clear and succinct visualization of the branched tests, each drilled out sample was filled with plaster (B08XW9ML9P, Amazon), treating the Sawbone sample as if it were a mold to be removed after hardening. The plaster model was then scanned into 3D CAD software (SolidWorks, Dassault Systèmes) using a laser 3D scanner (Artec Space Spider, Artec 3D Inc.). The created 3D model shown in Fig. \ref{fig:3D Branches} is the results of this process, and allowed us to quantitatively measure the trajectory's radius of curvature, length, and cut diameter in the laboratory without requiring complex imaging procedures such as a CT scan.

\vspace{-6 pt}
\section{Discussion}
The tests performed by the CT-SDR were designed to assess the repeatability and consistency of the drilled Trajectory 2 as prescribed by the biomechanics analyzed in section \ref{Traj Results} and then compared to the results discussed in previous works (i.e., \cite{alambeigi2019use,alambeigi2017curved,  ma2021active, wang2021design}). This was accomplished by analyzing the cross sections of the drilled pathways after experimentation (shown in Fig. \ref{fig:Sawbone Results}). The heat treated NiTi guide tube had an error between the actual shape and the biomechanically-obtained of 2.3\%. 
When drilling, every test had a standard deviation $<$ 0.75 mm, a minimum error of 1.7\%, and a maximum error of 2.2\% when compared to the actual shape of the nitinol guide. The highest error was recorded from the tests using the 0.85 mm/s insertion speed and 8250 rpm drill speed, and the lowest error was seen in the trials using the same drill speed but with an insertion of 1.25 mm/s.
When changing to the ball nose end mill, the trajectory errors did not shift from the range seen with the oval head bur. An average error of 3.0\% was measured. Differences in hole width was recorded between the two drill types, with the oval head having a hole width of 8.3 mm and the ball nose end mill having a hole width of 7.83 mm. 

In Fig. \ref{fig:3D Branches}, the surface of the drilling trajectories is best visualized. The smoothness can be both seen and felt from the plaster model, which demonstrates the limited vibrations in the system which would cause ridges and waves in the trajectories surface. Also shown in Fig. \ref{fig:3D Branches} are the radius measurements of the drilled trajectories from a shared entry point, these paths have an average radius of 70.8 mm, which reveals only a 2\% difference from the NiTi guide's actual radius. This result confirms the repeatability and consistency of the CT-SDR's ability to drill these curved trajectories both in and out of plane. 


\section{Conclusion and Future Work}
Toward reducing the failure rate of spinal fixation procedures in osteoporotic patients,  we proposed a novel biomechanically-aware design for a novel  CT-SDR. The proposed design leverages a patient-specific biomechanics model developed based on QCT scans of patient's vertebra to minimize the risk of a spinal fixation procedure by considering a biomechanically-optimal and feasible drilling trajectory.  Using the outputs of the FE framework, we designed  a robot that  can address the contradictory requirements of both the structural stiffness and adequate flexibility needed to deviate from the traditional linear pathways.  The FE biomechanical analysis showed that  utilization of flexible implants in an optimal curved trajectory can improve the stress and strains distribution in a patient's vertebra by 80.1\% and 77.5\%. Additionally, the results of testing the CT-SDR demonstrated a consistent drilling of both single and multiple drillings curved trajectories through Sawbone samples with an error between 0.14-4.1\% with respect to the planned trajectory by the NiTi guides. 
In the future, we will explore the biomechanics of more complex trajectories for flexible instruments implanted within a patient's vertebra. We will also consider biomechanics of implanted screws drilled via a CT-SDR with additional degrees of freedom \cite{sheela-CTSDR}. 

\vspace{-6 pt}

\bibliographystyle{IEEEtran}
\bibliography{ISMR2023_Handheld}

\end{document}